\documentclass[twocolumn,showpacs,amsmath,amssymb]{revtex4}
\usepackage{CJKutf8}
\usepackage{graphicx}
\usepackage{color}
\usepackage{booktabs}
\begin{document}
\title{Self-organizing Pattern in Multilayer Network for Words and Syllables}

\author{Li-Min Wang$^1$, Sun-Ting Tsai$^{1,\dagger}$, Shan-Jyun Wu$^1$, Meng-Xue Tsai$^1$,\\
	Daw-Wei Wang$^1$, Yi-Ching Su$^2$, and Tzay-Ming Hong$^{1,\ast}$\\
	$^1$Department of Physics, National Tsing Hua University, Hsinchu 30013, Taiwan, Republic of China\\
	$^2$Department of Linguistics, National Tsing Hua University, Hsinchu 30013, Taiwan, Republic of China\\
  \texttt{ming@phys.nthu.edu.tw} \\}

\date{\today}

\begin{abstract}
One of the ultimate goals for linguists is to find universal properties in human languages.  Although words are generally considered as representing arbitrary mappings between linguistic forms and meanings, we propose a new universal law that highlights the equally important role of syllables, which is complementary to Zipf’s. By plotting rank-rank (on frequency) distribution of (word, syllable) for English and Chinese corpora, visible lines appear and can be fit to a master curve. We discover the multilayer network for words and syllables based on this analysis exhibits the feature of self-organization which relies heavily on the inclusion of syllables and their connections. Analytic form for the scaling structure is derived and used to quantify how Internet slang becomes fashionable, which demonstrates its usefulness as a new tool to evolutionary linguistics.
\end{abstract}
\maketitle

\section{Introduction}
\subsection{Zipf's Law and Self-Organization}
Large data and prior knowledge are prerequisites in statistics, esp. in the field of computational linguistics\citep{ML in seg}. This reliance can be lessened considerably by the knowledge of universal laws that can help us categorize information\citep{cate}. A famous example is Zipf's law\citep{Zipf, critical review} which states that, given some corpus of natural language utterances, the frequency of word $N(x)$ and its rank $x$ in frequency-rank distribution (FRD) will exhibit a power-law relationship, $N(x)=ax^{-b}, $ where $b\sim 1$ varies with different languages\citep{different1, different2} and types of writing\citep{cate}. 

Self-organization is a dynamics process that describes the emergence of order without the control by an external agent. It appears in non-equilibrium systems\citep{soc}, living neurons\citep{s-o in neuron}, long-range temporal correlations in brains\citep{s-o in brain}, etc. In evolutionary linguistics, the existence of homophony reflects several self-organization characteristics\citep{s-o in language}. These features can be described by power laws when the process goes through a phase transition. For instance, the number of grammars
in terms of how many lexical categories and word order rules undergoes three emergent stages\citep{s-o in language}. However such properties are hard to find in existing data. In the following, we will provide a new method, rank-rank analysis, as a tool to reveal the secrets behind self-organization of languages.

\subsection{Statistical Properties Shared by Text Segmentation of English and Chinese}

Different languages have different grammatical rules to build sentences (e.g., SVO for English but SOV for Japanese) and words (e.g., affixation or compounding). We start by comparing two of the 
typologically remote languages, English (phonogram) and Chinese (logogram), and ask ``what is their common feature?". Noam Chomsky \citep{Chomsky} put forward universal grammar as one of the common features, which states that innate rules in children's brain are essential when learning a new language. The fact that young children with limited linguistic knowledge demonstrate adult-like knowledge of complex linguistic constraints supports such a proposal\citep{child1, child2}. In the mean time, NLP focuses on word-to-word relationships, e.g., context\citep{context} and collocation\citep{col0, col1}. All these informations can be used to construct word vector\citep{word2vec}. Unlike these perspectives which concentrate on ``word", we will show that the building block of words, i.e., syllables, also holds precious information on how languages develop. Evidence is provided that there exists a more fundamental statistical correlation between ``syllables" and words, that is common to Chinese and English.

\begin{figure}
	\centering 
	\includegraphics[width = 8cm]{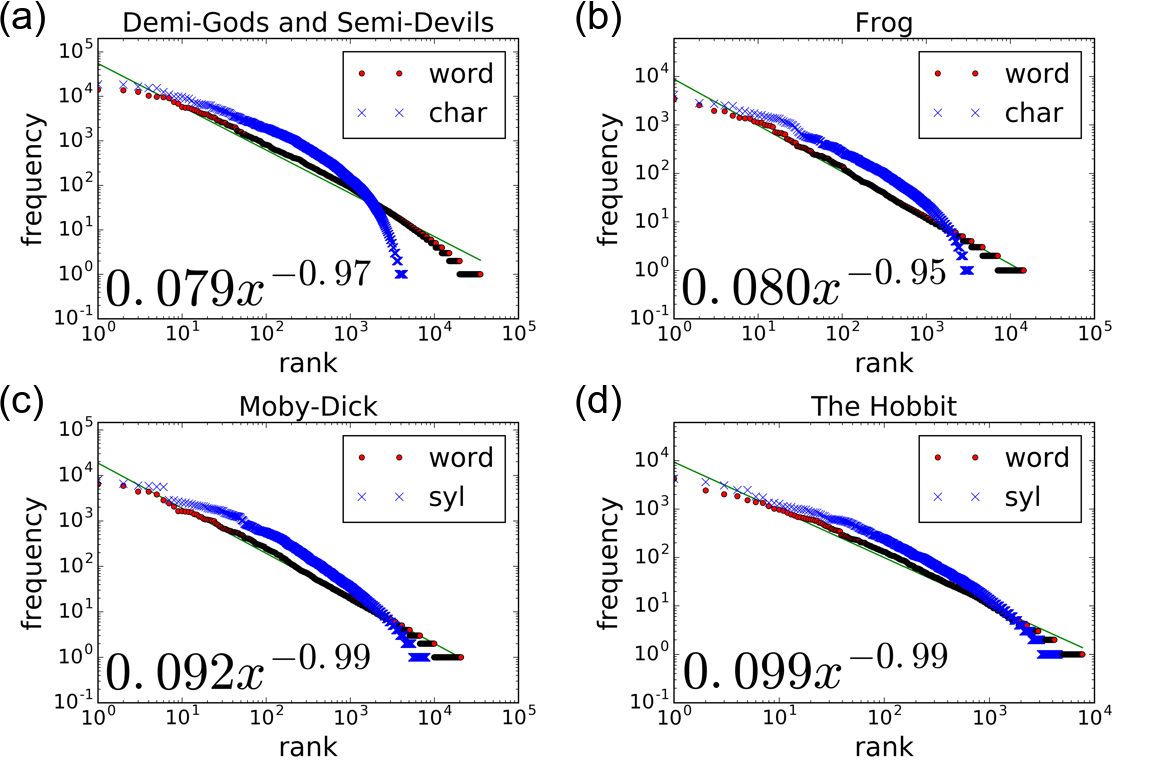}  
	\caption
	{The dotted/crossed (red/blue) curves in the log-log plots of (a, b) show FRD of word/syllable (character) in Chinese novels, {\it Demi-Gods and Semi-Devils} and {\it Frog}. Similarly, (c, d) are for word/syllable in English novels, {\it Moby-Dick} and {\it The Hobbit}. The fitting function is for the red points.
	}\label{fr-dis} 
\end{figure}

To begin with, let us provide some background knowledge. In Chinese texts, one syllable is written as one character which predominantly corresponds to a morpheme, and words are generally made up of multiple characters. To distinguish the typological difference between English and Chinese, we use character (char) to denote syllable for Chinese in the labels of the following figures. The first step to analyze Chinese text is to identify words from a chunk of text, which is more difficult than syllabifying words in English for the lack of symbols to segment word/syllable. Different segmentation algorithm will affect the statistical behavior of word vector. Without prior knowledge, Xiao\citep{Chinese Zipf} used $N$-gram to simulate Chinese words and concluded that Zipfian distribution exists for $N$=2, 3 and 4, but not 1.  
Apparently, this conclusion is not satisfactory because $N$ is not fixed, e.g.,\\

\begin{CJK}{UTF8}{bsmi}
	1-gram: 你、好、大、家、學、龍......
	
	2-gram: 家人、學校、生命、車子......
	
	3-gram: 語言學、選擇性、研究院......
\end{CJK}\\

With the aid of Sinica Corpus\citep{Sinica}, we are able to segment words precisely and verify the existence of Zipf’s law in Chinese by the red lines in Fig. \ref{fr-dis}(a, b). To be more rigorous in statistical analysis, the Akaike Information Criterion (AIC)\citep{AIC} has been employed to back up the claim. There has been confusion on whether Zipf's law remains valid when we switch to a smaller unit. The answer turns out to be negative for syllables in both Chinese and English, as shown by the blue curve in Fig. \ref{fr-dis}. How can we decipher the rule behind it and shed new light on Zipf's law?

\section{Self-Organizing Pattern in Multilayer Network for Words and Syllables}
\label{sec:network}
The hidden secret can be revealed by constructing rank-rank distribution (RRD). For Latin-based language/Chinese, let $(x, y)$=(rank of word, rank of syllable), where ranks depend on their frequency. Figure \ref{example}(a, b) exhibits the RRD in the Chinese/English novel, Frog/Moby-Dick, while \ref{example}(c, d) exemplify the process of plotting RRD. Ranks of word and syllable are decided by FRD as following:\\

$Rule\ 1$: {\it If several words or syllables share the same frequency, the one that appears earlier in the corpus will get a higher rank.}\\

To elucidate the linguistic and statistical reason behind the patterns visible from the graph, we build a multilayer network $G=(V,E,D)$ that contains two layers: word $G_w=$($V_w,E_w,w$)=(word, words sharing the same syllable) and syllable $G_s=$($V_s,E_s,s$)=(syllable, syllables sharing the same word), where $V\subset V_w\times V_s$ and $E\subset E_w\times E_s$. The vertices $(v^w_i, v^s_j)$ are mapped to $(x_i,y_j)$ with vertex weight $W_v(v^\alpha_i)=F(v^\alpha_i)$, where $F(v^\alpha_i)$ denotes the frequency of vertex $i$ in layer $\alpha$. The adjacency tensor $E^{(i,\alpha)}_{(j,\beta)}=\{e^{i\alpha}_{j\beta}\}$ can describe the edges between $v^\alpha_i$ to $v^\beta_j$ weights:
\begin{equation}
{e^{i\alpha}_{j\beta}}=
\begin{cases}
    F(v^\beta_j)-F(v^\alpha_i) &\text{if } \beta=\alpha \\
    0 \text{\ \ , if $i$ and $j$ are disconnected}\\
    1 &\text{otherwise}
\end{cases}
\end{equation}
where the first condition denotes the rank difference of two words/syllables, while the third condition represents the combination of syllables into a single word.

\begin{figure}[!t]
	\includegraphics[width = 7.5cm]{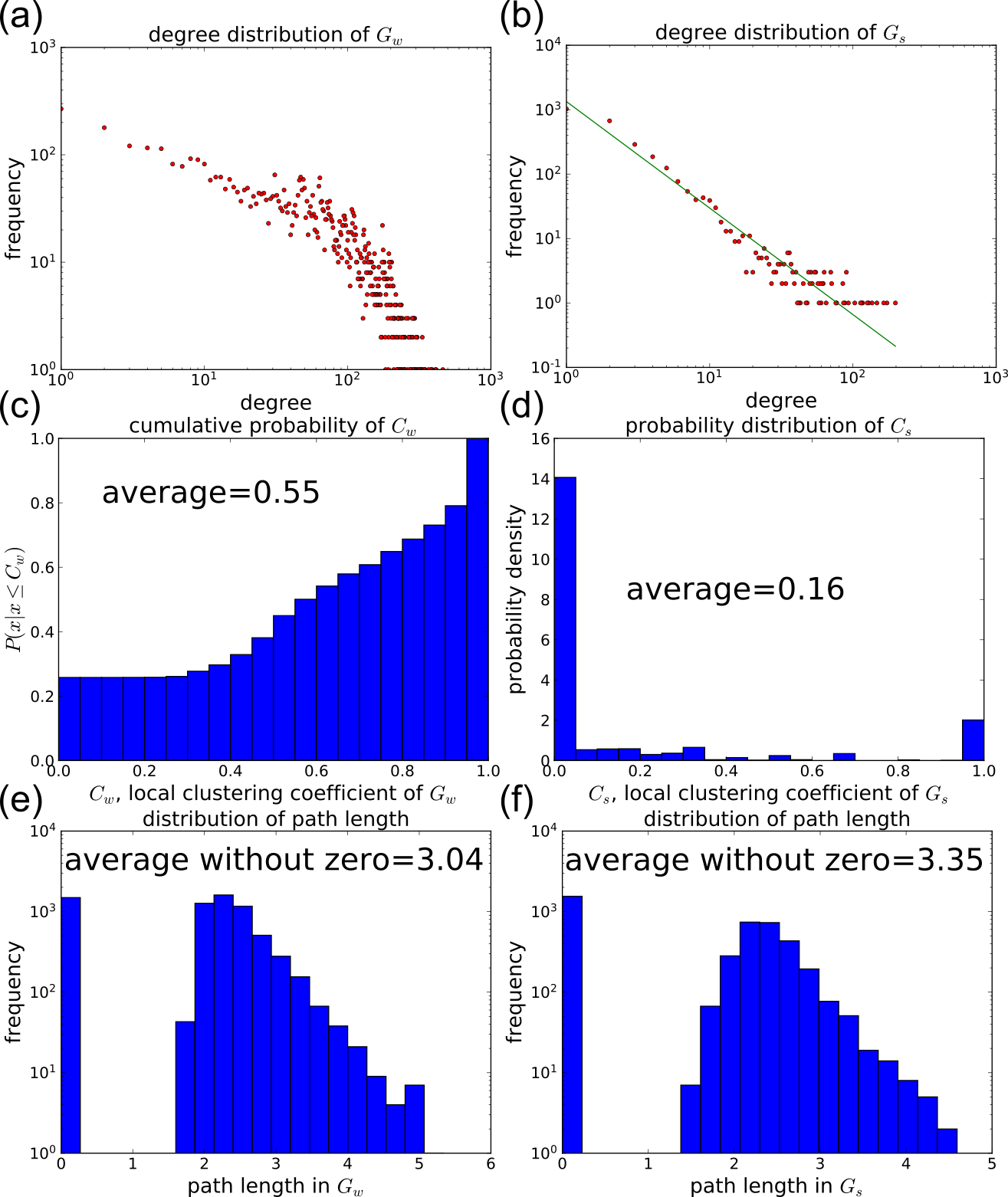}  
	\caption
	{The topological characteristics of multilayer network in Harry Potter 1. Panels (a, b) show the degree distribution of (word, syllable) where vertices with zero degree have been excluded because they present as a singularity. Note that (b) exhibits the feature of a scale-free network\citep{scale-free} $P(k_s)\sim k_s^{-\gamma}$. The distributions of local clustering coefficient of (word, syllable) are plotted in panels (c, d). The high average in (c) exemplifies a small-world network\citep{small-world}. As in (e, f), their average path length $(\bar{L}_w, \bar{L}_s)=(3.04,3.35)$.
	}\label{topo of network} 
\end{figure}

\begin{figure}[!h]
	\centering
	\includegraphics[width = 8cm]{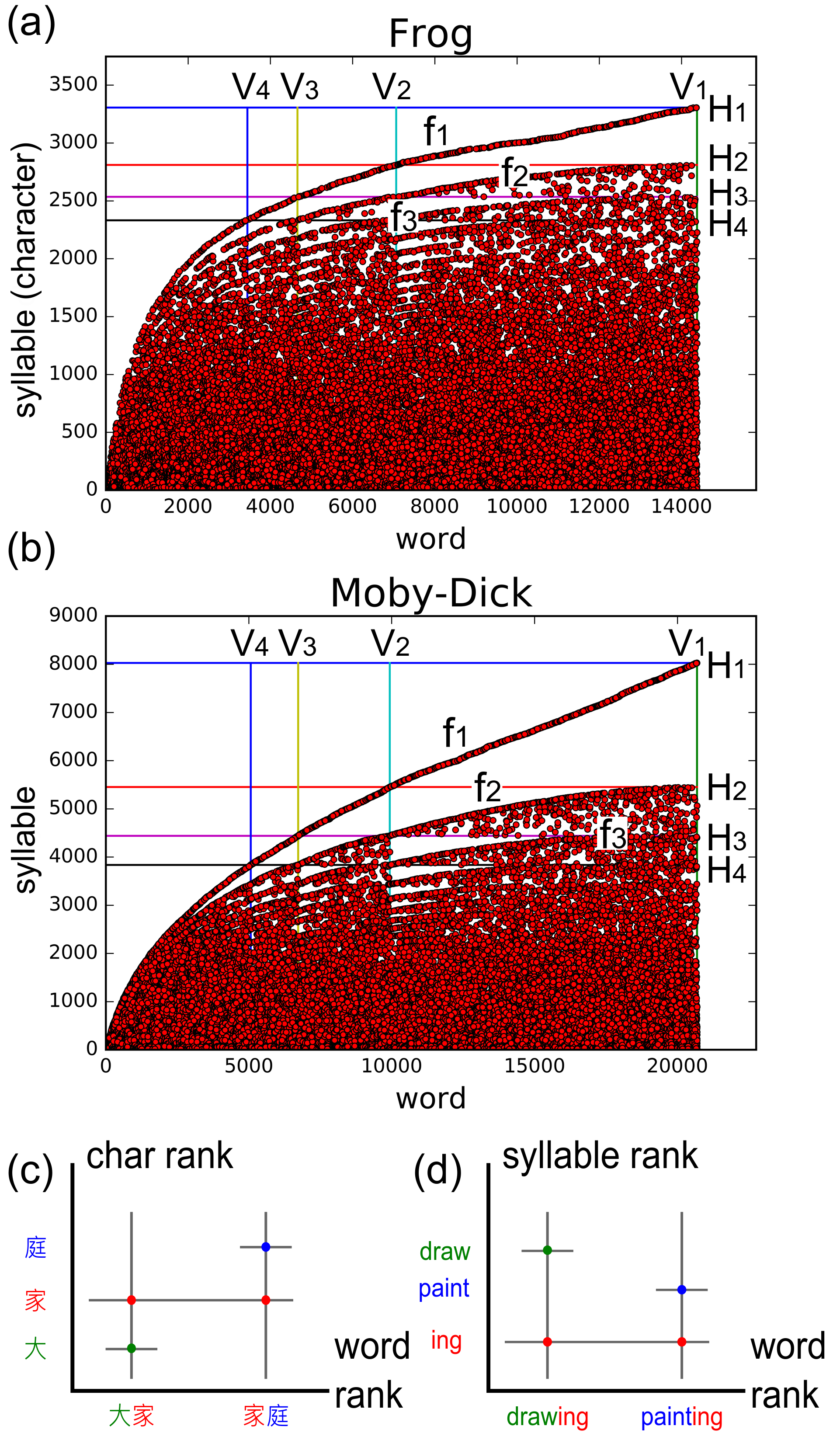}  
	\caption
	{ Panels (a, b) are the RRD plot for {\it Frog} and {\it Moby-Dick}.
		 Their construction is demonstrated schematically in panels (c, d).
	}\label{example} 
\end{figure}

There are two tools to analyze the RRD structure, that are based on either the multilayer network or the patterns. Let us start from the former by analyzing three topological properties that characterize the network: degree distribution ($P(k_w),P(k_s)$), clustering coefficients\citep{small-world} ($C_w, C_s$) and the shortest path length ($L_w, L_s$). To simplify our work, we assume all the nonzero weights to be unity. The degree distribution $P(k_\alpha)\equiv|\{v_{i}^{\alpha}\big\arrowvert\sum_j (w_e)^{i\alpha}_{j\alpha}=k_\alpha\}|$ where $|V|$ denotes the number of elements in set $V$.  The results are shown in Fig. \ref{topo of network}. A lot of isolated vertices are observed that do not connect to others. In the case of words, they refer to those whose syllables are never shared by other words. For syllables, they constitute single-syllable words. One should notice that the scale-free behavior is not based on FRD of syllables, but on the connection through words. We checked that the features shown in Fig. \ref{topo of network} are shared by other books which are listed in SM. Even we analyze the corpora mixed with different books, their power-law distributions still maintain. This is an evidence that ability to self-organize\citep{s-o in complex network} is manifested in the use of syllables. We will utilize this characteristics to develop a theory that explains how internet slang becomes fashionable in Sec. \ref{slang}.

Now we use the second tool, pattern, to understand RRD. The layer structure is clear to the eyes, which is verified by Fig. 3\&4 in Supplementary Material (SM) to obey a scaling relation upon scrutinization, where $r_f(x)\equiv{f_{n+1}(x)}/{f_{n}(x)}$,  $r_H\equiv{H_{n+1}}/{H_{n}}$ and $f_n (x)$ describes $n$-th layer in Fig. \ref{example}(a, b). Scaling implies $f_n (x)/f_m (x)$ is independent of $x$. 

What is amazing is that this {\bf scaling structure} (SS) not only appears in both English and Chinese corpora, but also in texts mixed with different styles of writing (see SM and No. 10 in Tab. \ref{t1}). This structure is another evidence that shows the feature of self-organization. The imminent question then is what causes this universal phenomenon. We were puzzled  for a long time until we add the horizontal $H_n$ and vertical $V_n$ lines which distinguish different groups that share the same frequency.\\

$Rule\ 2$: {\it The syllable within $(H_{n+1}, H_{n}]$ on RRD plot, where index $n$ labels the $n$-th line. For more than 90$\%$ data, those syllable exhibit the same frequency $=n$, except when $n$ is too large. Same for the words within $(V_{n+1}, V_{n}]$.}\\ 

Once these auxiliary lines are sketched, three useful observations appear. First, there is no ``fog" in the left region of $f_2$ where fog refers to the points not on any $f_i$. Second, each pair of horizontal and vertical lines {\it always} crosses the scaling lines. Third, the upper left area in each block is {\it always} devoid of fog.  \\

\section{Partition Theory of Scaling Structure}

Let's focus on $f_1$ which is unique because it lacks the fog that permeates the space between other neighboring layers. Since each syllable on $f_1$ has the same frequency as its associate word, it is impossible for this syllable to be shared by other words. For instance, {\it toms} in Long-bot-{\it toms}, {\it vior} in be-ha-{\it vior}, and  the 2-gram morpheme, \begin{CJK}{UTF8}{bsmi}尷尬 (embarrass) and 蹣跚 (stumble)\end{CJK}, in Chinese. This is why there is no point on the right side of $f_1$. The isolated vertices in multilayer network that we discussed in the last section fall straight on $f_1$.

Now let's turn our attention to the left side. Should there be any data point, it will imply its word frequency is bigger than that of its syllable - an apparent contradiction. This results in the vast blank space on the left side of $f_1$. To elaborate on this characteristics, we introduce the concept of ``partition function $p(N)$". The partitions of $N$ represent all possible permutations for syllable with frequency $=N$ to distribute over different words. For example, $p(3)=3$, where $N=3=2+1=1+1+1$, consists of three partitions. $N=3$ refers to a syllable that appears only in one word that in turn is used three times in the corpus. Similarly, $N=2+1$ means a frequency-$3$ syllable appears in two different words whose frequency $=2$ and $1$, respectively. Likewise, $N=1+1+1$ means three different words that all appear only once in the text. 

Under the framework of network, the above theory can be expressed by vertex weight function $w_v(v^\alpha_i)$. That is $w_v(v^s_i)=\sum_j w_v(v^w_j)$ where $v^w_j\in\{v^w_j\big\arrowvert v^w_j\in V_w, e^{jw}_{is}=1\}$, i.e., collection of words $\{v^w_j\}$ that contain syllables $v^s_i$. The RRD plot therefore can be decomposed into blocks, $\big\{(V_{m+1}, V_{m}], (H_{n+1}, H_{n}]\big\}$ labeled by $(m, n)$. The contour of scaling lines and the fog are both necessary components to the SS. 

According to $Rule$ 1, the earliest appearing syllable or word in each block must start from the left bottom corner. When later comers start to stack, they may go upward to form a line. Alternatively, they can move rightward and join the fog in the same or different blocks. This is why the three lines - horizontal, vertical, and scaling lines must cross. If the author uses any syllable in different words, it will create new points to the right of the earliest one. This explains why the upper left area in each block is guaranteed to be without fog. 

Nevertheless, such logical reasoning only gives partial story. Although we have checked that reversing $Rule\  1$, i.e., assign an earlier syllable or word a lower rank, would destroy the layer structure, we can not rule out that other variations may equally work. Let's try to understand the role of layer in each block. $Rule\ 1$ tells us that, whenever a word with new syllable appears in the block, the syllable should be placed at its top, i.e., becoming a member of the scaling curve. We confirm that  points from this mechanism constitute part of the layer structure.

\section{How Internet Slang Becomes Fashionable?}\label{slang}

To facilitate the identification of points on the scaling lines, knowledge of the analytic form of $f_n$ will help. Based on the scaling property, we assign $f_n(x)=a_nf(x)$. According to Fig. \ref{example}, 
\begin{equation}
H_n=a_mf(V_{n+1-m}).\end{equation}
By comparing Eq. (1) for each horizontal and vertical lines, we prove that
\begin{equation}
\frac{H_{n+1}}{H_{n}}=\frac{a_{n+1}}{a_{n}}=\frac{f(V_{n+1})}{f(V_{n})}=r\end{equation}
where ratio, $r$, is a constant of $n$. The evidence of this nontrivial consequence is presented in Fig. 3 in SM. We tend to use $r_f(x)$ rather than $r_H$ because the former covers more data. We discover from No. 12, 18 and 19 in Tab. \ref{t1} that: \\

$Rule\ 3$: {\it $\{H_n\}$ remains a geometric sequence even without the scaling structure.}\\

In other words, the existence of a constant $r_H$  is likely a mere statistical outcome. By use of Zipf's law and $Rule\ 2$, $ n\approx N(\frac{V_{n+1}+V_{n}}{2})=a(\frac{V_{n+1}+V_{n}}{2})^{-b}$. Since the mid-point of each block $(\bar{x},\bar{y})=(\frac{V_{n+1}+V_{n}}{2},\frac{H_{n+1}+H_{n}}{2})$ roughly falls on the scaling line, it can be shown that 
\begin{equation}
\bar{y}\approx a_1f(\bar{x})\approx H_{1}(1+\frac{1}{r})r^{a\bar{x}^{-b}}.
\end{equation}
Note that Eq. (3) can be used to derive the form of scaling function as 
\begin{equation}
y\approx Ar^{N(x)}\label{theory of scaling}
\end{equation}
where $N(x)$ is FRD of word. The SS and Eq. (\ref{theory of scaling}) quantify the hidden rule of word-composition based on the usage rate of syllables. This can be observed from the differential form of Eq. (\ref{theory of scaling}), ${dN}\propto {dy/y},$ that reveals the increase of word usage ($dN$) is proportional to the difference in popularity among syllables ($dy$) weighted by the inverse of their rank ($1/y$), similar to the rich-get-richer spirit of the preferential attachment\citep{barabasi} that leads to scale-free behavior. As shown in SM, the scaling function can be used to recognize scaling patterns; therefore, expand the method of text categorization\citep{cate} by pattern recognition.

The above conclusion can be generalized to explain and quantify the fashionable usage of Internet slang like ``LOL" (Laughing Out Loud) and ``AFK'' (Away From Keyboard) where L, O, F, and K become new members of syllable. Thus these alphabets are sure to fall on the scaling lines according to $Rule\ 1$, and follow Eq. (\ref{theory of scaling}). This explains why the usage of their compound words\citep{compound} increases so fast once these simple and trendy syllable enjoy a high rank $y$. From the above analysis, we realize that scaling lines can function as a filter for new words.

Several techniques, such as standard error optimization and denoising, are used to determine the parameters in Eq. (\ref{theory of scaling}) and other important quantities: goodness of fitting ($G$) and scaling ($SP$), number of total words ($L$), and size of word bank ($V_1$), defined as the upper bound of word rank $x$.  Details can be found in SM.

\section{Analysis of Linguistic Factors}

Equation \ref{theory of scaling}, strictly speaking, is a phenomenological theory and key to understanding SS. Let us address some questions to gain more insights: Is SS a consequence of Zipf's law (question 1)? What is the role of  $L$ (question 2)? Is the grammar also a deciding factor (question 3)? Is SS common to different styles of writing (question 4)? We will thus check poems and proses in addition to novels. Another legitimate enquiry is whether a collection of works by a single author or various authors, such as newspaper, still retain SS (question 5).

\begin{table*}[!h]
	\caption{
		Statistical quantities of different corpora. All corpora obey Zipf's law, except No. 16$\sim$19. No. $6\sim 8$ are scientific articles in English. Data for 60 more corpora can be found in Supplementary information.}\label{t1} 
	\begin{tabular}{cccccc}
		\toprule
		No. & Sample        & Zipf ($a, b$)   & Scaling ($SP, M, V_1, L$)&Language\\ 
		\midrule
		1&{\it Moby-Dick}	& (0.092, 0.99)   & (0.905, 2402, 20688, 203k)     	&English	\\ 
		2&{\it The Hobbit}	& (0.099, 0.99)   & (0.871, 642, 7690, 94k)       &English	\\ 
		3&Xu Zhimo \begin{CJK}{UTF8}{bsmi}徐志摩\end{CJK} 			& (0.078, 0.86)   & (0.713, 36, 1235, 3k)    	&Chinese	\\ 
		4&{\it Frog}	\begin{CJK}{UTF8}{bsmi}蛙\end{CJK}	& (0.080, 0.95)   & (0.862, 682, 14380, 110k)		&Chinese	\\
		5&{\it Demi-Gods and Semi-Devils}	\begin{CJK}{UTF8}{bsmi}天龍八部\end{CJK}	& (0.080, 0.97)   & (0.88, 1632, 35223, 695k)		&Chinese	\\
		6&LIGO 2016	 		& (0.053, 0.81)   & (0.734, 336, 2830, 7.9k) 		&Sci. English 	\\ 
		7&Chopstick  		& (0.072, 0.84)   & (0.637, 144, 1153, 3.1k)    	&Sci. English	\\
		8&Empirical Test of Zipf  		& (0.083, 0.86)   & (0.345, 95, 910, 2.8k)    	&Sci. English	\\
		9&Newspaper	  			& (0.032, 0.74)   & (0.869, 150, 4143, 12k)      	&Chinese	\\
		10&Mix	(various authors)        	& (0.079, 0.96)   & (0.893, 981, 22716, 174k)      	&Chinese	\\ 
		11&Paper generator\cite{paperG}  & (0.068, 0.85)   & (0.138, 344, 2128, 26k) 	     &Fake English	\\ 
		12&1-gram Fake      & (0.108, 0.95)   & (NAN, 1, 1541, 20k)    	&Fake Chinese	\\ 
		13&2-gram Fake      & (0.097, 0.96)   & (0.719, 105, 4019, 20k)    	&Fake Chinese\\ 
		14&3-gram Fake      & (0.097, 0.96)   & (0.849, 157, 3931, 20k)   	&Fake Chinese	\\ 
		15&4-gram Fake      & (0.097, 0.96)   & (0.779, 265, 3964, 20k)    	&Fake Chinese  \\ 
		16&2-gram log-normal    & (0.096, 0.96)	   & (0.756, 181, 4146, 20k)    	&Fake Chinese  \\ 
		17&2-gram double power law 		& (0.333, 1.40)	   & (0.762, 68, 3308, 200k)    	&Fake Chinese  \\ 
		18&2-gram exponential   & (0.049, 0.69)	   & (0.039, 17, 596, 20k)    	&Fake Chinese  \\ 
		19&2-gram Gaussian  & (0.005, 0.42)	   & (0.537, 105, 3561, 20k)    	&Fake Chinese  \\ 
		20&Excerpts from Frog	    & (0.051, 0.71)	   & (0.423, 33, 637, 1.2k)    	& Chinese  \\
		21&Excerpts from Moby-Dick  & (0.064, 0.77)	   & (0.318, 58, 653, 1.4k)    	& English  \\
		\bottomrule
	\end{tabular}
\end{table*}

To clarify these questions, we summarize statistical quantities of different corpora in Table \ref{t1} and organize our analyses in Table \ref{t2}. First, the answer to question 1 is negative. The proof is provided by two examples: (1). An extreme article, No.12, consists of 1-gram words whose FRD follows Zipf's law, but its RRD is a straight line, i.e., no SS. (2). Article No. 16 and 17 obey Zipf-like FRD, but still show SS. Second, SS is not guaranteed by a large $L$ because No. 18 and 19 show no SS, while No. 13$\sim$15 with the same article size (about 20000 words) do. However, a small $L$ is sure to ruin SS, as evidenced by No. 20 and 21. Third, in response to question 3, No. 11, a man-made corpora based on grammar, has a very low $SP$. In contrast, No. 12$\sim$17 that do not follow rules of writing acquire high $SP$. They indicate that grammar and real words are inessential. Fourth, the answer to questions 4 and 5 is a sound yes, as proven by the high $SP$ for No. 9 and 10. This implies\\

$Rule\ 4$: {\it Mixing different literary styles will not ruin the scaling structure.}\\

\begin{table*}[!h]
	\caption{
This table determines that a sound SS, represented by a large $SP$ value, relies on a large size of word bank $V_1$, and can still exist for fake corpora that consist of words composed of random characters or do not follow rules of writing.
	}
	\begin{tabular}{|c|c|c|c|c|}
		\hline
		& corpora 	& scientific 			& fake, No.11			& fake, No.12$\sim$17  \\
		&			& article				& (real words)& (random words)\\ \hline
		real words 	& yes 		& yes 					& yes 			& no \\ \hline
		grammar 	& yes 		& yes 					& yes			& no \\ \hline
		$V_1>1200$ 	& yes 	& no 				& yes 		& yes \\ \hline
		Zipf or 	& yes 	& yes 				& no 		& yes \\
		Zipf-like&			& 				& & \\ \hline
		$SP>0.75$	& yes 		& no 					& no 			& yes \\ \hline
	\end{tabular}
	\label{t2}
\end{table*}

Last, but not the least, candidate to affect SS is $V_1$. In No. 3 (poems), 6, 7, and 8 (scientific articles) whose $SP$ falls below 0.75, which prove that a small word bank is detrimental to SS.

\begin{figure}[!h]
	\centering
	\includegraphics[width=8cm]{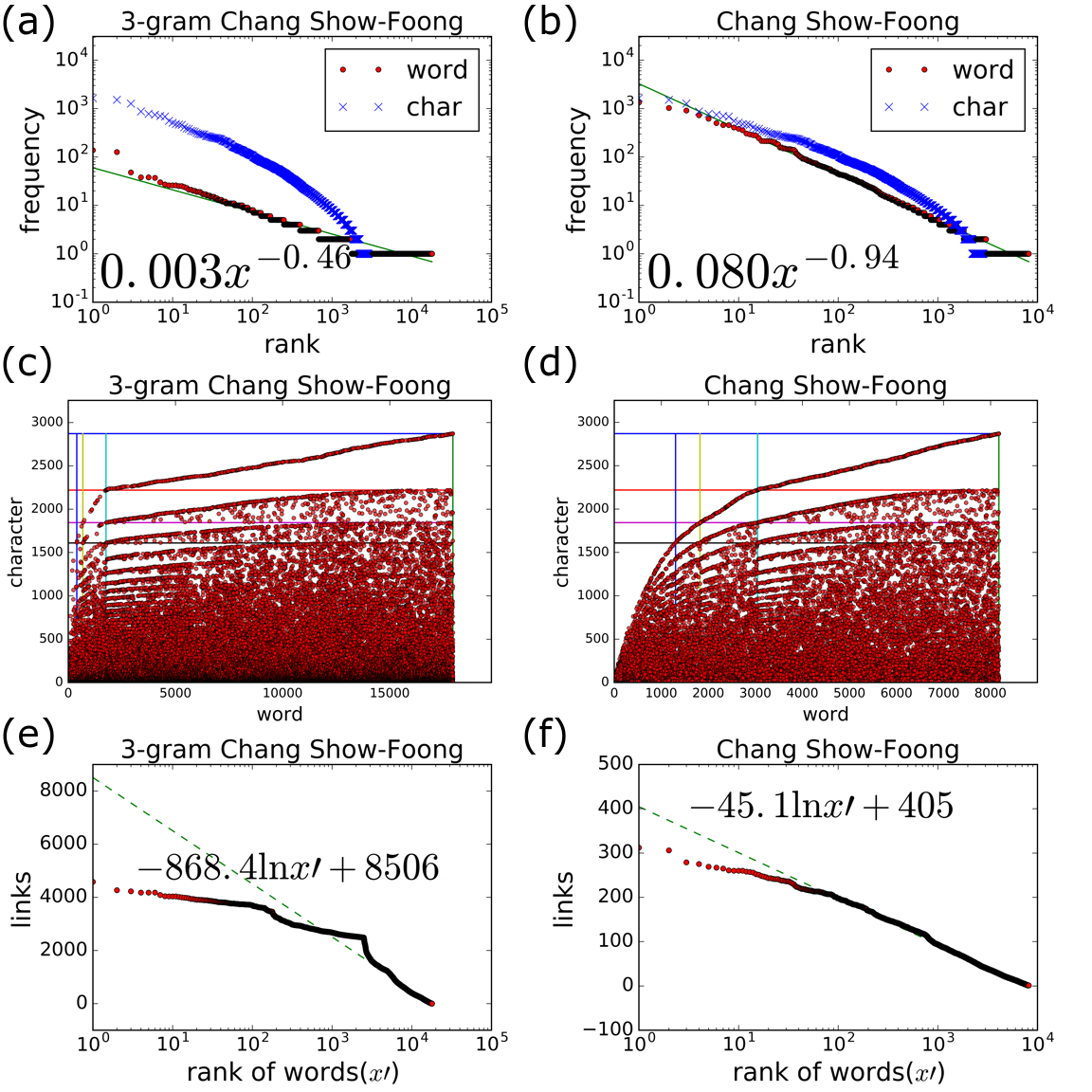}  
	\caption
	{ For Chang Show-Foong's book left figures exhibit 3-gram version, while the right ones use Academia Sinica word segmentation\citep{Sinica}. Panels (a, b) are the FRD plot, (c, d) RRD plot, and (e, f) $Link$-rank plot. The most significant difference between these two versions is NOT in their FRD, but the smoothness of RRD and $Link$ function. Since Sinica version is better developed than 3-gram and other N-gram which also exhibit unsmooth points, we can conclude that 
the smoothness  of RRD and $Link$ function is an ideal statistical standard to judge the maturity of an algorithm for word segmentation.}\label{Ngram}
\end{figure}

Now we shift our focus to the elements of word bank. Remembering the logical reasoning in the paragraph following $Rule\ 2$, the partitions not only result in several blocks, but also impose a constraint on how words are composed.  We analyze the morphology of words by generalizing the concept of collocation\cite{col0, col1} to define two functions: $Col(v^s_i):V_s\to R$, which stands for the number of different words that contain the syllable $v^s_i$, and  $Link(v^w_j):V_w\to R$ as

\begin{equation}
\begin{aligned}
&Col(v^s_i)\equiv \big\arrowvert\big\{v\big\arrowvert v^s = v^s_i,\forall v=(v^w,v^s)\in V\big\}\big\arrowvert\\
&Link(v^w_j)\equiv \sum_i Col(v^s_i)
\end{aligned}
\end{equation}
where $v^s_i$ in $Link$ follows $v^s_i\in\big\{v^s\big\arrowvert v^w = v^w_j,\forall (v^w,v^s)\in V\big\}$. Elements in $\{\}$ is non-repeat.

For instance, if syllable A appears in words AB, AC, and KAD,  then $Col(A)=3$; if syllable H appears only in H or HH, then $Col(H)=1$; if syllable T appears in both T and TT, then $Col(T)=2$. $Link(ABC)=Col(A)+Col(B)+Col(C)$ and $Link(AA)=Col(A)$. We observe a useful parameter to gauge the popularity of word is 
\begin{equation}
Link(x)\leq M
\end{equation}
where $M$ represents the biggest $Link$ in the word layer of network.

In Sec. \ref{sec:network}, the function form of degree distribution of word is undetermined. With the aid of $Link$ and $Col$ functions, a simple relation can be revealed in Fig. 5 in SM. These two functions are useful at characterizing RRD with a faster computational speed. Before discussing how these two functions affect SS, let us introduce word segmentation as an application of $Link$ function. 

\section{Application of $Link$: Gauge the Accuracy of Text Segmentation}

Thousands of new words appear yearly in the Internet community. To analyze their statistics, we need to deal with word segmentation first. Figure \ref{Ngram} shows two different algorithms. From their FRD, it is hard to tell their difference. But on the other hand, we can easily rank their supremacy by the smoothness of RRD and $Link$. Most importantly, these properties can be used to reinforce the N-gram-based classification\citep{cate} in addition to Zipf's law. At the same time, RRD and $Link$ also offer common characteristics on tokenization\citep{phonetic} and word segmentation\citep{word seg} in different natural languages.

\begin{figure}[!h]
	\centering
	\includegraphics[width=8cm]{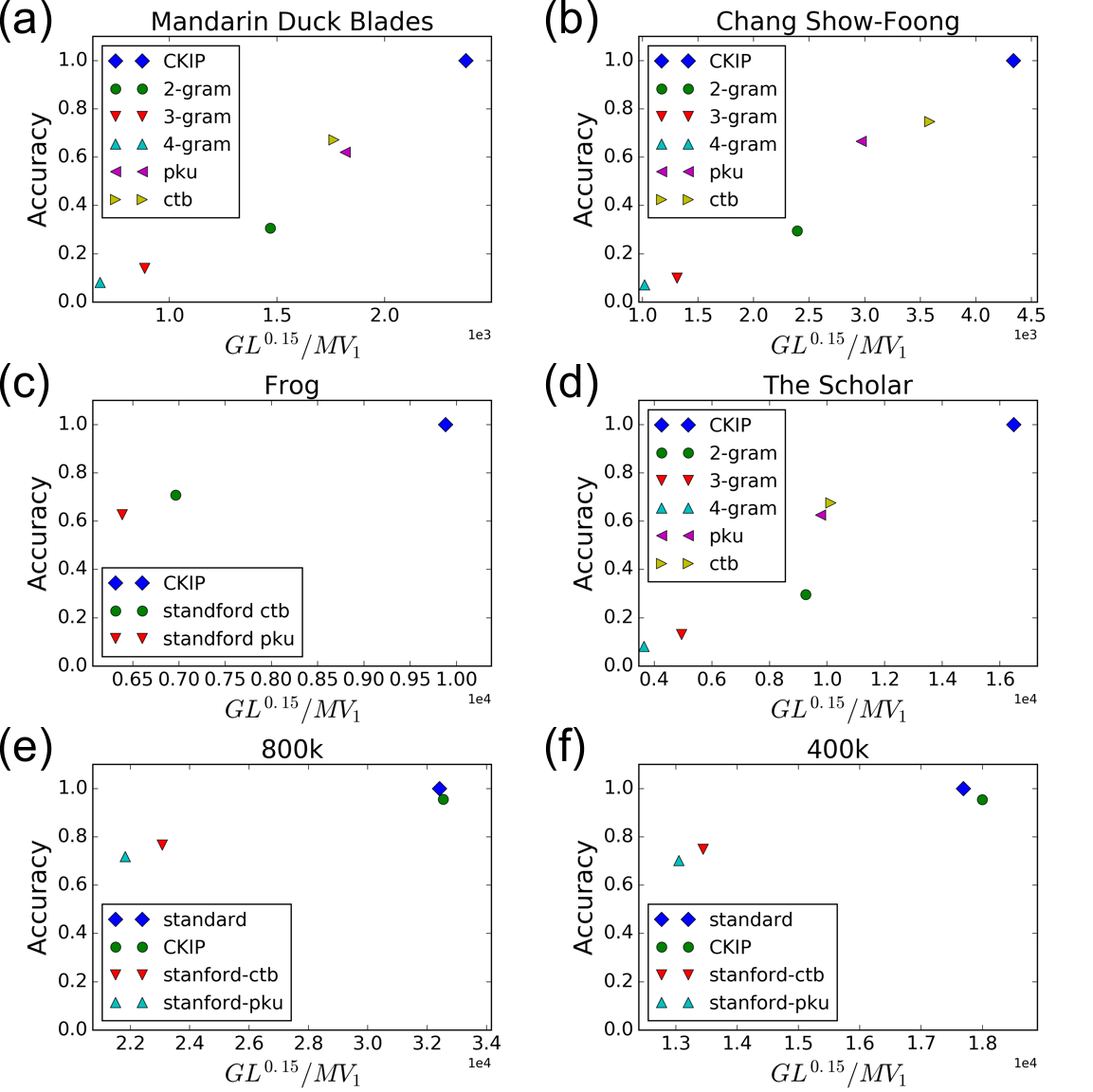}  
	\caption
	{ The $A-GL^{0.15}/(MV_1)$ plot for corpora segmented by different algorithms\citep{standford1, standford2}. We use CKIP\citep{Sinica}, the best Chinese algorithm, as the standard  in cases (a, b, c, d), while (e, f) employ the word bank\citep{standard bank} segmented by linguists in Academia Sinica. Initially we were alarmed by the violation of positive correlation in (e, f). But soon it dawned on us that the version of standard bank we used is older than the one by CKIP. This realization reveals that the $A-GL^{0.15}/(MV_1)$ plot has the potential to function as an empirical index for the accuracy of algorithm.
	}\label{Accuracy}
\end{figure}

How to gauge the goodness of segmentation algorithm in practice is our next task. Traditionally it is determined by accuracy based on a standard segmented corpus by human or statistical observations like word frequency. Alternatively, similar effects can be achieved by studying the properties of RRD plot, such as $G$, $M$, $L$, and $V_1$. We define accuracy $A$ as the ratio of numbers of ``correct'' syllables/words in certain algorithm and total syllables/words in a standard segmented English/Chinese corpus.  ``Correct" means the syllable/word can be found in a standard segmented corpus. The validity of our new method is supported by Fig. \ref{Accuracy} which proves that $A$ and $GL^{0.15}/(MV_1)$ are positively correlated.  

The index $GL^{0.15}/(MV_1)$ is significant at presenting a graphical method.  It becomes handy when cracking an unknown ancient language or that of animals whose data and knowledge are limited. For this study, we build N-gram Fake texts, e.g., No. 12$\sim$19 in Tab. \ref{t1}, whose words are composed of N random\citep{Miller} characters to simulate an unidentified linguistic rule. The original corpora contain spaces; act as a segmentation symbol between N-gram words. We define words in original corpora as standard words and run different algorithms to segment them in spaceless corpora. It is expected that most available algorithms are doomed  because they are tailored for existing languages, but RRD and $Link$ can help decide the supremacy among newly developed algorithms.

\section{Connection to Zipf's and Heaps' Laws}

Before concluding, we want to investigate the statistical behavior of $Link(x')$ and  $Col(y')$ where $x'$ and $y'$ denote the new rank of word in $Link$ and syllable in $Col$, respectively. Figure 5 in SM suggests the form of $Link$ as:
\begin{equation}
Link(x')=-C\ln{x'}+D\label{Link function}
\end{equation}
and $Col$ as:
\begin{equation}
Col(y')=(-E\ln{y'}+F)^2\label{Col function}
\end{equation}

From the definition of these two functions, we can deduce that $D/C\approx\ln{V_1}$ and $F/E\approx\ln{H_1}$ by taking $x'=V_1$ and $y'=H_1$, where $V_1$ and $H_1$ denote the size of word and syllable banks.

Now let us clarify the role of length of utterances, $L=\int_{1}^{V_1}{N(x)dx}$ where $N(x)$ is FRD of word. In addition to Zipf's law, there exists a second empirical law by Heaps\citep{heaps}, which states that $V_1=\kappa L^{\beta}$ where $\kappa=10\sim 100$ and $\beta=0.4\sim 0.6$ in English text corpora. After combining it with Eqs.(\ref{Link function}, \ref{Col function}), we obtain 
\begin{equation}
M=Link(1)=C[\beta\ln(L/1)+\kappa]
\end{equation}
that integrates Zipf's and Heaps' laws, and SS we discovered.

\section{Conclusion}
We have introduced SS and $Link$ as new fundamental properties of statistical linguistics, in complement to Zipf's law. (i). Besides word-to-word relationship, syllable is shown to be equally crucial in language evolution because word-composition exhibits the feature of self-organization, which is evidenced by its scale-free network.  (ii). They are not only new empirical laws, but also a new approach to transforming linguistic problems into graphical ones.  (iii). The indices, $SP$ and $GL^{0.15}/(MV_1)$, are important in this research. The former describes the goodness of scaling, while the latter endows SS a practical usage in computational linguistics. (iv). We suggest an index $GL^{0.15}/(MV_1)$ to gauge the accuracy of segmentation algorithm. (v). The analytic form of the scaling function in Eq. (\ref{theory of scaling}) has been proven useful at understanding the popularity of Internet slang. We expect that it will be equally fruitful to apply our RRD analysis to evolutionary linguistics.

\section{Future Work}
The properties of SS may be used to study languages uttered by babies, non-humans, minorities, etc., which may suffer the deficiency of limited data and knowledge. Future work should focus on two subjects: further studies on the mechanism behind SS and a more comprehensive index than $SP$. The latter arises because of two reasons: Although cases in Fig. 6 in SM have been judged as $SP<0.75$, (i). Figure 6(a, b), being excerpts from natural corpora, ought to exhibit SS. (ii). The different RRD patterns of fake SCI paper in Fig. 6(e) and the real one in Fig. 6(f) are obvious. Will the implementation of SS to computer-generated texts makes it more like those by real human being? As the great Tang poet and painter, Wang Wei, was famous for: ``There is poetry in his painting and painting in his poetry", it is reasonable to expect our concept of RRD will have impact beyond statistical linguistics and bring new ideas to other art forms, such as painting, music\citep{Zipf in music}, and sculpture. This generalization may bring useful statistical laws as soon as the correct unit and its elementary unit are identified. We can verify the relationship between  units and sub-units by RRD, which is expected to exhibit SS. New statistical laws are useful at improving the flexibility of AI,  which is guided mainly by rule-based approaches so far.

\end{document}